\documentclass[review]{elsarticle}

\usepackage{hyperref}
\usepackage{multirow}
\usepackage{xcolor}
\usepackage{ulem}
\usepackage{float}
\usepackage{graphicx} 
\usepackage{subcaption} 
\usepackage{makecell}

\journal{Journal of \LaTeX\ Templates}

\begin{document}
	
	\begin{frontmatter}
		
		\title{Speeding up the annotation process in semantic segmentation industrial applications} 

		\author[dasci]{Marta Fernández-Moreno\corref{mycorrespondingauthor}}
		\ead{martafdezm95@gmail.com}
		
		\author[ingenica]{Margarita Guerrero}
		\ead{marga.guerrero@ingenicasts.es}
		
		\author[arcelor]{Rosalia Rementeria}
		\ead{rosalia.rementeria@arcelormittal.com}
		
		\author[dasci]{Pablo Mesejo}
		\ead{pmesejo@decsai.ugr.es}
		
		\author[uned]{Raúl Moreno}
		\ead{raul.moreno.salinas@gmail.com}

		\address[dasci]{Department of Computer Science and Artificial Intelligence, Andalusian Research Institute in Data Science and Computational Intelligence, DaSCI, University of Granada, 18071, Granada, Spain}
		\address[ingenica]{Ingénica STS, Av. Pablo Iglesias 38, Bj,  33205, Gijón, Spain}
		\address[arcelor]{ArcelorMittal Global R\&D,  SLab—Steel Labs, Calle Marineros 4, 33490, Avilés, Spain}
		\address[uned]{Department of Computer Science and Automatic Control, National Distance Education University (UNED), Juan del Rosal 16, Madrid 28040, Spain}
		\cortext[mycorrespondingauthor]{Corresponding author}
		
		\begin{abstract}
		Current machine learning models commonly require large and well-annotated datasets. However, the annotation process often becomes a bottleneck, with increased complexity leading to higher chances of human errors. Within this context, our goal in this paper is to leverage unsupervised algorithms to improve data annotation efficiency for complex semantic segmentation problems in industrial materials science. Previous research has quantified labeling time and others explored unsupervised methods. However, to the best of our knowledge, this is the first study to quantify how much unsupervised algorithms accelerate the labeling process. We aim to validate the extent to which this laborious process can be accelerated, focusing on semantic segmentation tasks that involve annotating each pixel of high-resolution images, such as the microstructure characterization challenge in materials science. Specifically, we demonstrate that by using unsupervised computer vision algorithms, the time required for the labeling process can be reduced from 170 hours to 37 hours, achieving an approximate reduction of 78\%. The dataset we work with includes large images of dimensions 1280x959 and 960x703, which further increases the complexity of the annotation task. Despite these challenges, we create and share the largest public steel microstructure segmentation dataset to date, available under MIT License with permanent DOI, contributing a fully annotated, high-resolution dataset to the field. Additionally, this is the first work to compare the labeling time from scratch (a common approach in previous studies) to the labeling time when using these unsupervised algorithms as a pre-annotation step. Furthermore, we provide a Deep Learning model trained on this dataset, validated by field experts, and deployed in an industrial setting, serving as an initial benchmark for this public dataset.
		
		\end{abstract}
		\begin{keyword}
			Semantic Segmentation \sep Steel Microstructure Dataset \sep AI-Powered Data labeling System \sep Industrial applications \sep Unsupervised learning  \sep Deep convolutional neural networks
		\end{keyword}
		
	\end{frontmatter}
	
	
	\section{Introduction}
	
	The advances in the field of machine learning (ML) over the last few years have increased notably its presence in different fields such as industry \cite{ge2017industry}, material science \cite{Luengo2022microstructures}, healthcare \cite{ferdous2020healthcare} and economics \cite{athey2018economics}. Data quality plays a key role in these applications as it can affect predictions such as disease detection \cite{ferdous2020healthcare}. However, more attention is often paid to algorithmic tasks and model optimization, leaving data acquisition and data quality in the background. After interviewing more than 50 data scientists, the authors of \cite{Sambasivan2021DataWork} have identified the various negative effects and problems triggered by putting data quality behind. One of these problems is the huge amount of time spent on fine-tuning a model based on incorrect data that might be the root of the poor performance of the model. Researching and enhancing models is essential, but data is equally crucial. Generating a dataset while tackling an ML project, requires selecting meaningful samples and generating clear labels for the given task. However, the labeling process is often a tedious and time-consuming task, which leads to the introduction of human errors. 
	Previous research as \cite{northcutt2021labelling} has shown that mislabeling in datasets, including popular computer vision (CV) datasets like ImageNet, can contribute significantly to model inefficiencies, finding out that approximately \begin{math}3.3\%\end{math} of the labels are erroneous, rising to \begin{math}6\%\end{math} in cases such as ImageNet \cite{deng2009imagenet}.
	
	Moreover, semantic segmentation is a CV task where the label of each image is manually annotated pixel by pixel, turning it into one of the most effort-intensive labeling tasks. This can be seen in public datasets such as MS COCO \cite{lin2014coco}, where image-level labeling for classification tasks took an average of 4.1 seconds, while pixel-level labeling for semantic segmentation tasks took 10.1 minutes. Depending on the labeling taxonomy and the nature of the images, the labeling time can vary greatly. One example is the Camvid dataset \cite{gabriel2009camvid}, where images count with a high number of classes, so the labeling of each image took 60 minutes on average. Another example is Cityscapes \cite{cortds2016cityscapes}, a public dataset with high-resolution images (1024$\times$2048). These studies clearly indicate that the annotation process is a major bottleneck in the development of reliable datasets for machine learning.
	
	At higher resolutions, the annotation effort increases due to the finer granularity of the labeling unit, which is the pixel. This higher resolution introduces additional complexities in the labeling task, including difficulties in identifying objects in challenging areas, even for an expert annotator, as well as dealing with a larger number of objects and ensuring the continuity of labels across adjacent pixels. Moreover, the level of detail required in the annotations, such as precise contours or identification of very small yet important sections, increases the complexity of the task. Due to this, labeling a Cityscapes image takes 90 minutes on average. MetalDAM \cite{Luengo2022microstructures}, requiring ~140 min/image due to its complexity, serves as reference for MicroSteel's challenges (Section 3.2).
	
	Currently, there are unsupervised methods \cite{ji2019unsupervised,kanezaki2018} and foundation models that, while not precise enough as a standalone solution, are accurate enough to serve as an initial label for general tasks. However, they cannot be directly applied for highly specific tasks requiring greater precision and real-world industrial applications. In the case of the segment-anything (SAM) model, the authors acknowledge its general nature and note that SAM may not achieve optimal results in specific domains, often being outperformed by other approaches \cite{kirillov2023sam}. Previous research on annotation-efficient semantic segmentation has relied on active-learning clicks \cite{ge2024esa,cai2021revisiting} and weak or coarse supervision \cite{li2019weaklier,jing2019coarse}; nevertheless, all those approaches still require some form of human interaction or image-level labels before the first mask is produced.

	In the case of semantic segmentation problems, these methods usually perform well in segmenting majority classes, requiring only annotating minor classes and fixing very few errors. In the case of microstructure characterization datasets, both \cite{Luengo2022microstructures} and the one presented in this paper contain two highly unstructured majority classes with very complex boundaries. The present paper demonstrates that, based on these datasets, the performance of unsupervised deep learning segmentation algorithms is sufficiently strong to be valuable in a subsequent phase of label refinement conducted by a human.
	
	
	In this paper, we address the first study that explicitly compares labeling time from scratch to labeling time when using pre-annotations generated by unsupervised algorithms. By comparing these two approaches, we quantify the significant efficiency gained in the annotation process. In practical terms, material characterization often requires human experts to manually label microstructure images, a process that is both time-consuming and error-prone. Quantifying these classes by hand introduces variability and inefficiency, which motivates the need for automating this procedure. To fully automate the material characterization task, it is necessary to train a robust model. However, the challenge lies in the need for large annotated datasets to train such a model, making the annotation process a key bottleneck. This paper proposes a methodology that accelerates the labeling process, ultimately facilitating the creation of a model capable of automating the entire task.
	
	We empirically prove how much this labeling process can be accelerated by applying unsupervised algorithms for image segmentation. Since our interest is to study to what extent labeling time can be reduced, we choose a complex and challenging problem whose datasets require very time-consuming manual annotation. For this reason, we start from a problem similar to the one presented by MetalDAM (segmentation of microstructures in metallography)\cite{Luengo2022microstructures}. Then, we proceed as follows:
	
	\begin{enumerate}
		\item First, we select an industrial problem in material science where a certain material needs to be characterized. Therefore, images of steel microstructures are collected using microscopes by material science experts.
		\item As a first procedure, every pixel in the images is labeled by experts.
		\item As a second procedure, some labels are also generated by using CV algorithms with a low computational cost for training and inference. These labels are later refined by a material science expert.
		\item Finally, the labeling time and the quality of the labels are compared between both procedures (manual vs semi-automatic).
	\end{enumerate}
	
	In addition, this work presents a procedure for solving a computer vision use case in a quick manner by making use of CV algorithms in both the dataset generation process and use case resolution.
	
In summary, the main contributions of this article are:
\begin{itemize}
	\item We demonstrate how unsupervised machine learning algorithms can reduce the manual annotation time for real-world problems by approximately \begin{math}78\%\end{math}, decreasing the effort from 170 hours to just 37 hours.
	\item Following the previously mentioned unsupervised methodology, we create the largest publicly available semantic segmentation dataset of steel microstructures to date, consisting of 82 fully-annotated high-resolution images (1280$\times$959 and 960$\times$703). The dataset will be made available at https://github.com/martafdezmAM/microsteel.git.
	\item We apply the two aforementioned contributions to solve a real-world industrial use case in material science, showing how the proposed methodology can significantly improve efficiency in automatic material characterization.
\end{itemize}
	

\section{Proposed Methodology}
The main goal of this paper is to provide a deep study and an optimization of one of the main bottlenecks in ML projects: the process of generating a reliable labeled and curated dataset. To do so, we employ as a benchmark a semantic segmentation problem due to the high complexity of its pixel-wise labels. MicroSteel exhibits the annotation challenges detailed in Section 3.2. Every dataset usually contains majority classes, so in the case of highly unbalanced datasets like this one, automating the labeling of majority classes helps to drastically speed up the annotation process.

\begin{figure}
	\caption{\textbf{Comparison of Manual Labeling and Pre-Annotation-Based Labeling Strategies:} This figure illustrates the two labeling strategies compared in this study.The first labeling strategy is described in the upper part of the image, which involves manual labeling, where an expert annotates each image from scratch. The second labeling strategy is therefore described at the bottom of the image, where pre-annotations generated by unsupervised algorithms are then refined by the expert to produce the final annotations.}
	\label{fig:labelling_strategies }
	\centering
	\includegraphics[width=12cm]{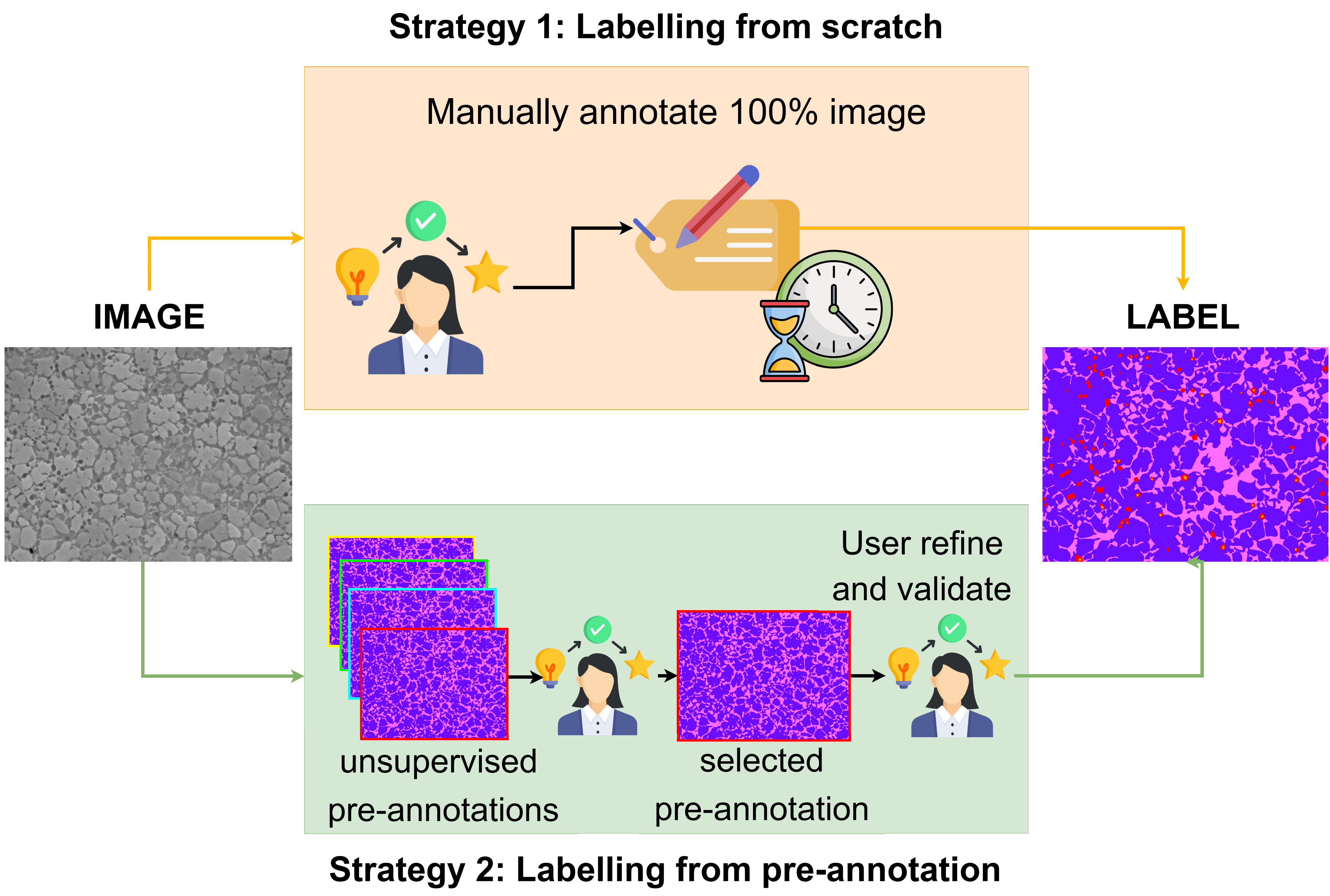}
\end{figure}

In this study, two labeling strategies are compared: labeling from scratch and labeling from pre-annotations. As shown in Figure \ref{fig:labelling_strategies }, in the first strategy, a materials science expert labels each image from scratch to produce the final label. In the second strategy, the expert is provided with an initial pre-annotation, which is modified to generate the final annotation. For the second strategy, pre-annotations are generated using various image segmentation techniques, and these pre-annotations are loaded into a standard annotation tool such as Labelbox \footnote{Labeling tool: \url{https://labelbox.com/}}, where experts can refine them. The goal is to maximize the similarity of pre-annotations to the final labels while minimizing the time needed for manual correction.

The annotation process consisted in organizing and uploading all images and pre-annotations to Labelbox. This platform automatically recorded the precise time spent on each annotation task, from initial loading to final submission, for both strategies. We calculated median annotation times across experts to compare the efficiency of both approaches while accounting for inter-expert variability.

The pre-annotation process involves the following key considerations:

i) Unsupervised semantic segmentation algorithms are used to automatically label each pixel without prior information.
ii) Simple, fast algorithms are prioritized to avoid training or execution processes that would take hours or days.

Several methods have been evaluated, such as \cite{ji2019unsupervised,cho2021unsupervised,hwang2019unsupervised}, but due to their reliance on deep learning models requiring extensive training, they were deemed unsuitable for pre-annotation purposes. Instead, the following simpler techniques were selected:

\begin{itemize}{
		\item \textbf{Thresholding}: We apply the multi-Otsu thresholding method \cite{Liao2001multiotsu}, an enhancement of Otsu's threshold \cite{otsu1979threshold}, which is based on grey-level histograms.
		\item \textbf{Superpixels}: The graph-based superpixel segmentation algorithm \cite{felzenszwalb2004superpixels} is used, which segments the image based on local neighborhoods.
		\item \textbf{Clustering}: The k-means clustering algorithm \cite{hartigan1979kmeans} is applied, treating each pixel as a data point.
		\item \textbf{DL method}: A fully unsupervised convolutional neural network proposed by \cite{kanezaki2018} is used. This method \textbf{trains from scratch on each individual image} using a loss function that jointly optimizes feature similarity, spatial continuity, and the number of unique clusters $K$. No pre-training or labeled data is required. This approach has been validated on similar microstructure characterization datasets in \cite{kim2020metallurgic} and \cite{fernandez2023tradeoff}.
		\item \textbf{Segment Anything (SAM)}: We also include \textbf{Segment Anything (SAM)}, a state-of-the-art segmentation model developed by Meta AI \cite{kirillov2023sam}. SAM is a general-purpose segmentation model that generates high-quality masks for a wide range of objects. For our dataset, SAM was used to generate preliminary masks, which were then post-processed to merge overlapping regions and extract majority classes.
}\end{itemize}

The hardware (Intel Core i9-11900K, 126 GB RAM, NVIDIA RTX 3090) used classical methods such as multi-Otsu and superpixels run in sub-second time per image, while k-means typically takes a few seconds per image. GPU based methods range from sub-second to a few seconds for SAM inference, and tens of seconds for the DL-based method due to per-image optimization.

\begin{table}[h]
	\centering
	\caption{Computational cost of pre-annotation methods measured on Intel Core i9-11900K and NVIDIA RTX 3090.}
	\label{tab:runtimes}
	\begin{tabular}{lcc}
		\hline
		\textbf{Method} & \textbf{Runtime per image} & \textbf{Hardware req.} \\
		\hline
		Multi-Otsu & $<$1 s & CPU (Intel Core i9-11900K) \\
		Superpixels & $<$1 s & CPU (Intel Core i9-11900K) \\
		K-means & 2--5 s & CPU (Intel Core i9-11900K) \\
		SAM & 1--3 s & GPU (NVIDIA RTX 3090) \\
		DL-based & 10--20 s & GPU (NVIDIA RTX 3090) \\
		\hline
	\end{tabular}
\end{table}

Table \ref{tab:runtimes} summarizes the computational cost of each pre-annotation technique. Classical methods (multi-Otsu, superpixels, k-means) run efficiently on CPU and complete pre-annotation for the entire dataset (82 images) in 1-6 minutes. GPU-based methods (SAM, DL-based) require dedicated hardware but remain practical for industrial workflows: SAM processes the full dataset in approximately 4 minutes, while the DL-based method requires approximately 20 minutes. These pre-annotation times (ranging from minutes to less than one hour) are negligible compared to the 37–170 hours required for manual annotation, confirming that computational overhead does not limit the practical applicability of our approach.

The DL-based pre-annotation method was trained independently for each image using a fixed and small number of optimization epochs (typically 100–150 epochs), following the original implementation in \cite{kanezaki2018}, which is designed for fast convergence without requiring large-scale training.

As the pre-annotation process takes place before the dataset is fully labeled, a qualitative evaluation was performed by an expert in the field to select the best mask for each image from the five techniques. The unsupervised DL-based algorithm and SAM were consistently among the top performers, with SAM excelling in images where objects were well-defined and not highly intermixed (e.g., Type I), while our DL-based method showed superior performance in handling complex boundaries and intermixed classes (e.g., Type II and Type III). The selected pre-annotations were then loaded into annotation tools for refinement by the expert.

As shown in Figure \ref{fig:labelling_strategies }, the final labels are generated by minor adjustments to these pre-annotations, resulting in two distinct labels per image, one from scratch and one from pre-annotations. Both the labeling quality and time will be compared between these strategies later in this study.
	
\section{MicroSteel dataset}
This section introduces the data that was generated, utilized, and shared in this study, as well as the problem that this type of dataset aims to solve. As mentioned before, one of the contributions of this paper is the generation of a fully annotated industrial dataset for steel microstructure characterization. This dataset was efficiently produced using unsupervised segmentation methods. 
The MicroSteel dataset is publicly available at \href{https://github.com/martafdezmAM/microsteel}{github.com/martafdezmAM/microsteel} (MIT License) and archived on Zenodo \href{https://doi.org/10.5281/zenodo.18826160}{(DOI: 10.5281/zenodo.18826160)} for permanent preservation.

\subsection{Microstructure characterization}
In materials science the size, amount, distribution, and morphology of the microstructure constituents are studied in the literature, frequently to find correlations with the mechanical properties of the materials \cite{reed1973physical}. This analysis depends on either manual measurement of optical microscopy images or automated measurements of electron backscatter diffraction scans. However, these techniques are limited to simple and coarse microstructures and are also time-consuming for materials science experts. In recent years, semantic segmentation techniques have become very popular in this field, they can provide automatically all this information of great interest to experts \cite{holm2020overview}. 

Table \ref{tab:datasets} summarizes the most relevant information from the existing material characterization datasets. As can be seen, \cite{Luengo2022microstructures} and \cite{DeCost2019HighTQ} are the only annotated datasets which are public. As mentioned in \cite{Luengo2022microstructures}, labeling an image with these characteristics takes an average of 4 hours, which complicates the generation of such datasets, so speeding up this process is crucial. Moreover, making data publicly available is essential to enable the potential of AI and it is also an important aspect of materials science, as this data is made available to organizations specialising in different fields of research for their usage.

\begin{table}[!htp]\centering
	\caption{Summary of the currently available metallography datasets \cite{Luengo2022microstructures}}\label{tab:datasets}
	\resizebox{\textwidth}{!}{
		\begin{tabular}{lcccc}
			\hline
			\textbf{Dataset} &\textbf{Images} &\textbf{Resolution} &\textbf{labeled} &\textbf{Availability} \\
			\hline\noalign{\smallskip}
			\citet{zhang2019aluminum} &5086 &2560$\times$1920 &no &private \\
			UHCSDB~\cite{decost2017UHCS} &961 &645$\times$484 &no &public \\
			\citet{Roberts2019Deep} &2 &2048$\times$2048 &yes &private \\
			\citet{zhang2019aluminum} (subset) \cite{li2020online} \cite{Chen2021Semi}&30 &1024$\times$768 &yes &private \\
			UHCSDB (subset)~\cite{DeCost2019HighTQ} &24 &645$\times$484 &yes &public \\
			MetalDAM~\cite{Luengo2022microstructures} &42 &1280$\times$895, 1024$\times$703 &yes &public \\
			MetalDAM (unlabeled)~\cite{Luengo2022microstructures}&164&1280$\times$895, 1024$\times$703&no&public\\
			\hline
	\end{tabular}}
\end{table}

\subsection{MicroSteel}
MicroSteel is a new image segmentation benchmark in material science. It consists of 82 images with resolutions 1280$\times$959 and 960$\times$703, annotated at the pixel level. Compared with datasets in Table \ref{tab:datasets}, MicroSteel not only is fully labeled and publicly accessible, but also has a much higher volume of fully annotated images than any other labeled dataset in this field (public or private). This makes MicroSteel the largest public semantic segmentation dataset of microstructures available to the best of the authors' knowledge.

\begin{figure}[H]
	\caption{Example micrographs and corresponding pixel-wise masks from the MicroSteel dataset for the three industrial processing routes (Type I, Type II and Type III) in steel microstructure characterization. Each row shows an original optical micrograph (left) and its final expert-generated mask (right), used as ground truth for training and evaluation. Each class represents a distinct phase in the steel microstructure, color-coded as follows: Alpha (index 0, purple), TiB2 (index 1, red), TiN (index 2, yellow), FeTiB (index 3, green) and Fe2B (index 4, pink)}
	\label{fig:dataset_samples}
	\centering
	a) Type I \\
	\includegraphics[width=5.5cm,height=4cm]{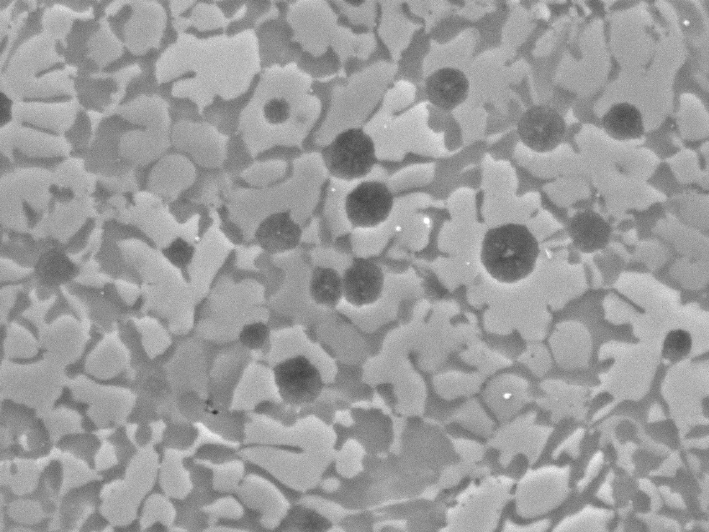}
	\includegraphics[width=5.5cm,height=4cm]{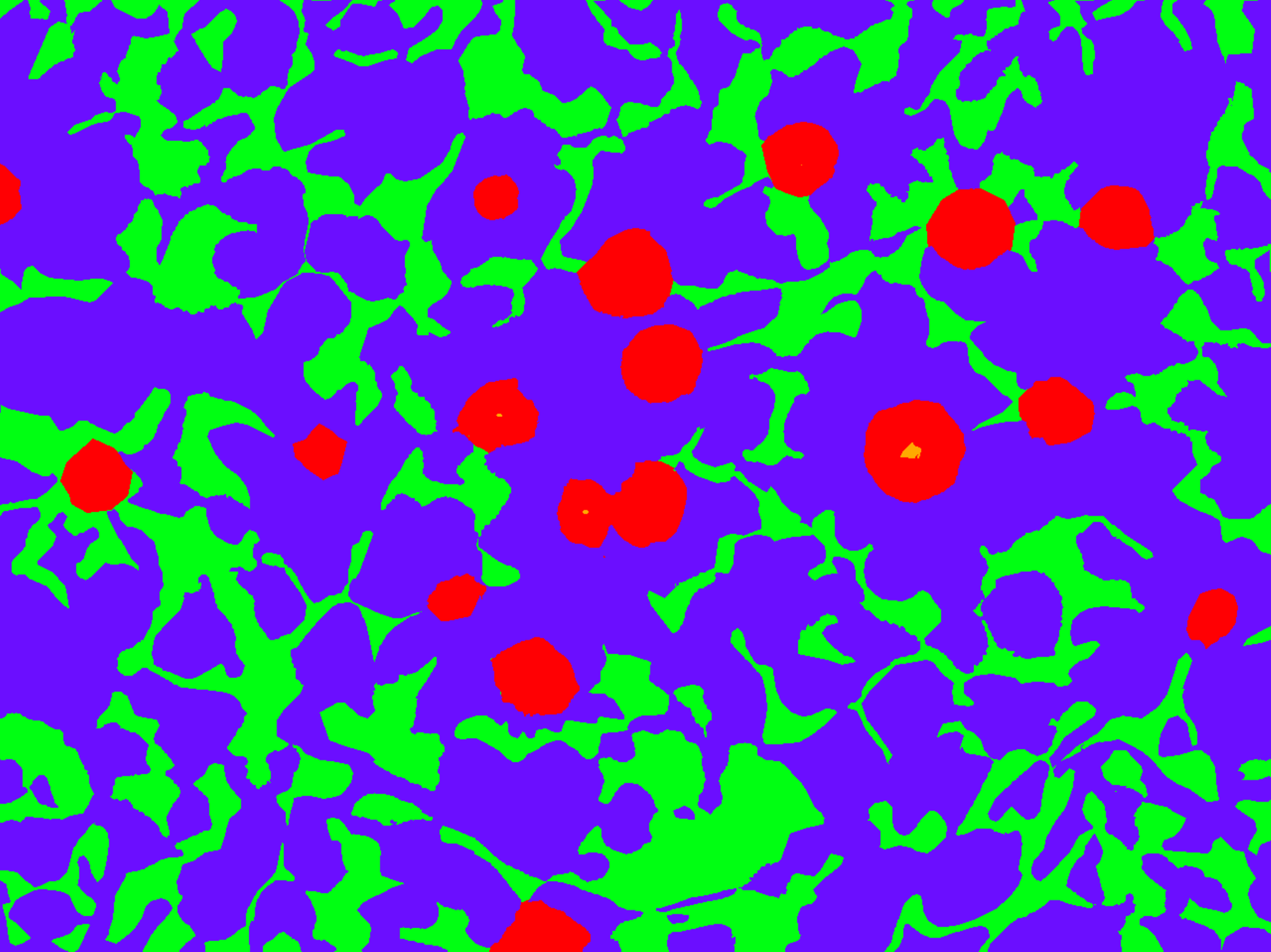}
	\\ b) Type II\\
	\includegraphics[width=5.5cm,height=4cm]{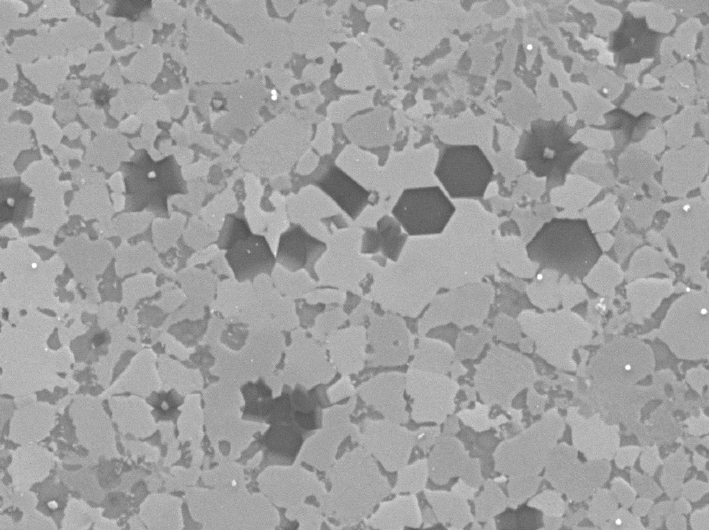}
	\includegraphics[width=5.5cm,height=4cm]{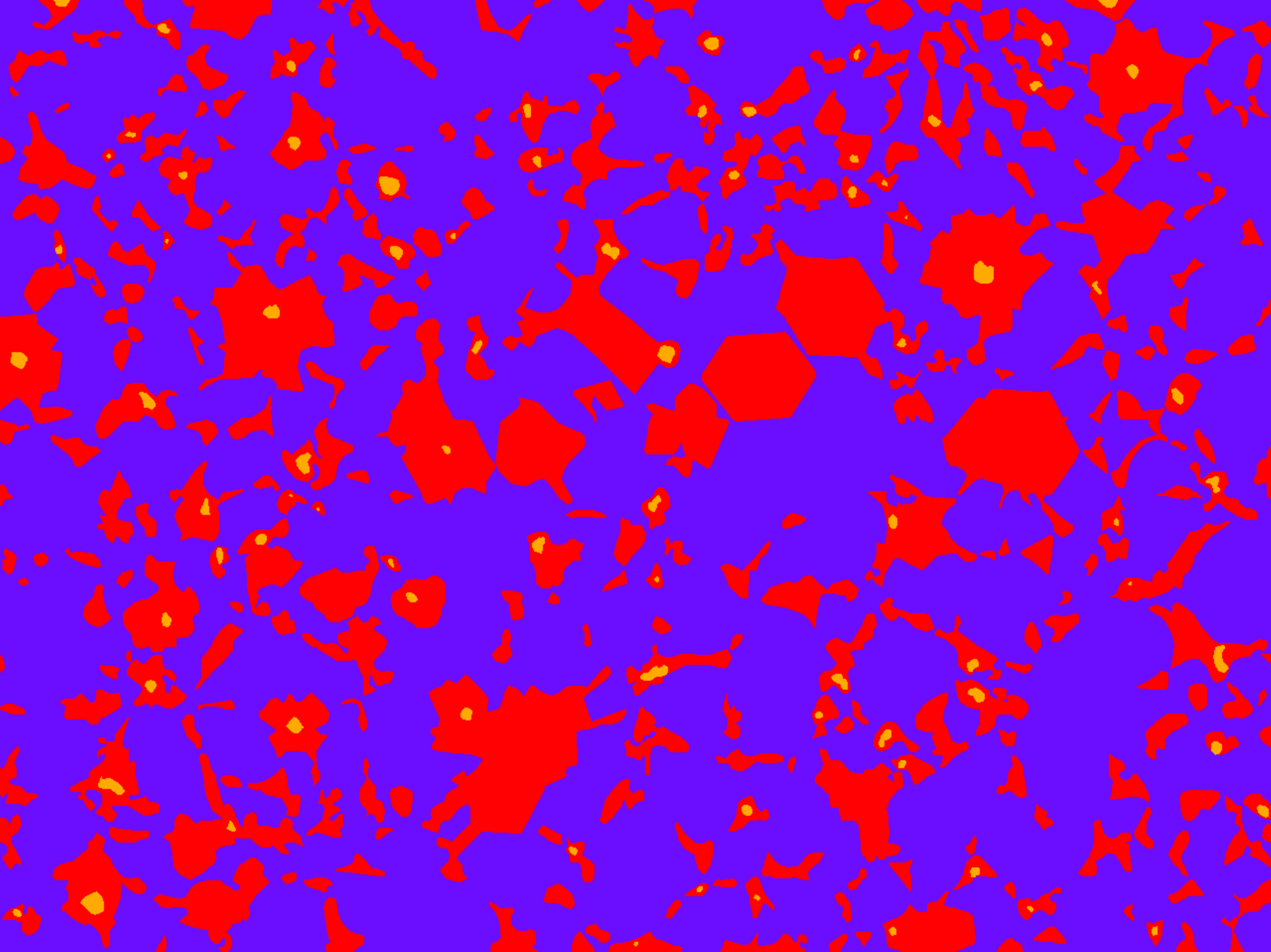}
	\\ c) Type III\\
	\includegraphics[width=5.5cm,height=4cm]{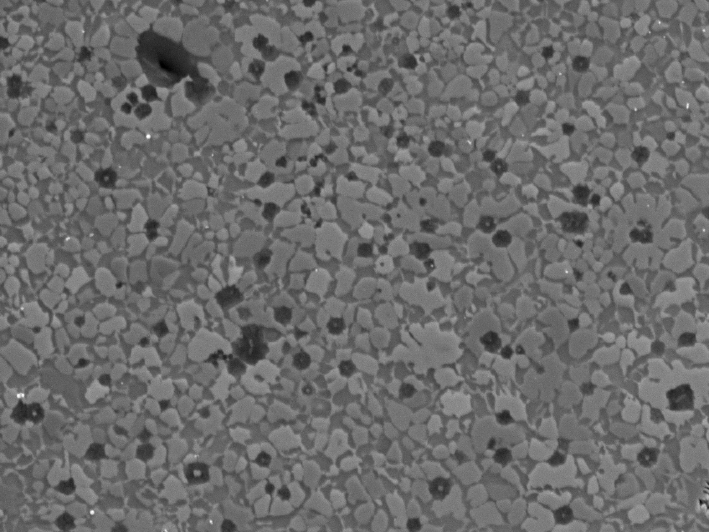}
	\includegraphics[width=5.5cm,height=4cm]{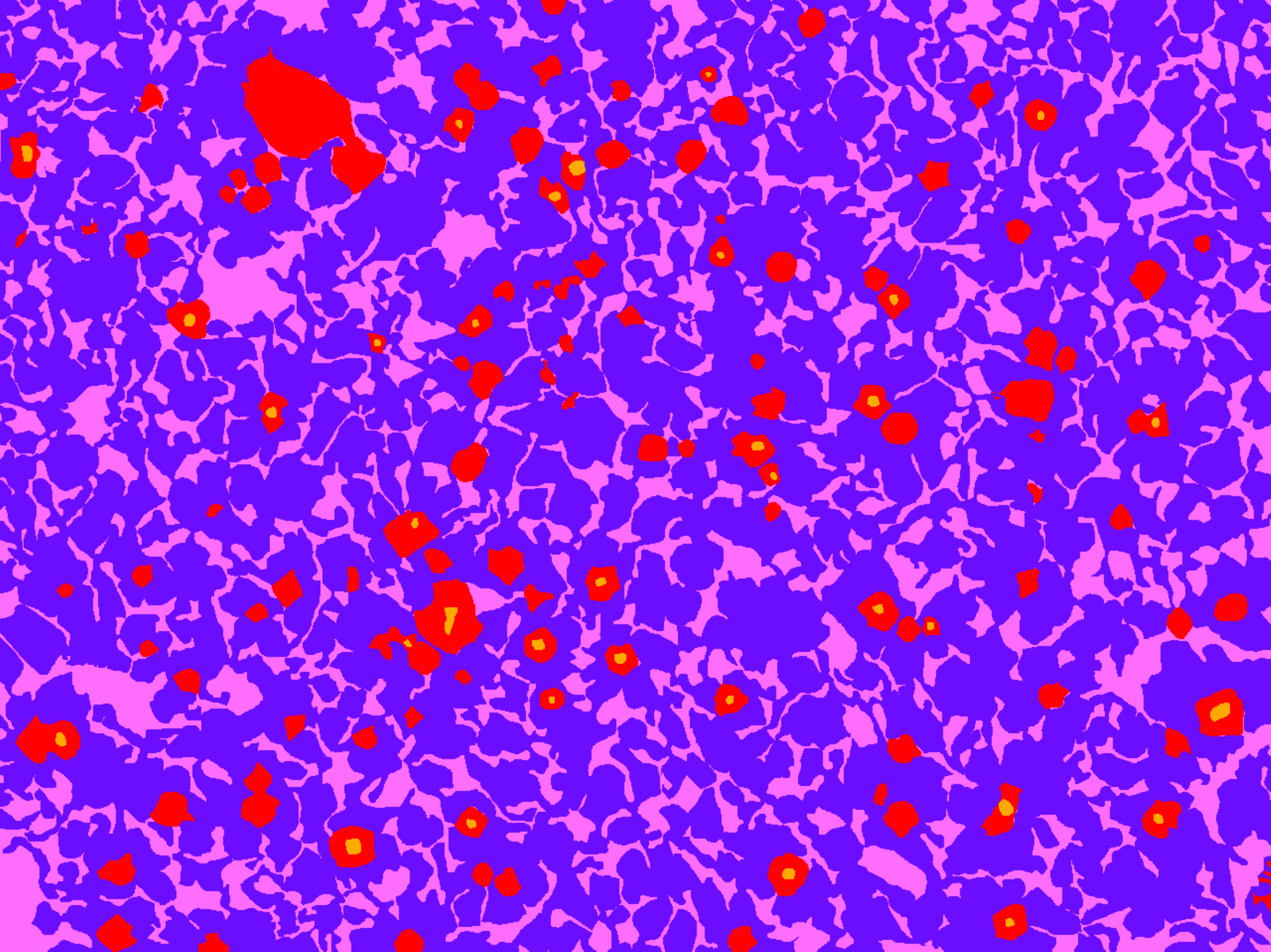}
\end{figure}

In this dataset, 5 different classes have been identified, related to each of the microconstituents to be analysed, and are represented by a different color in Figure \ref{fig:dataset_samples}. In addition, we differentiate the dataset into 3 types according to the type of industrial processing each material has undergone, Figure \ref{fig:dataset_samples} shows an example of an image and its mask for each type. From the knowledge of experts in materials science we know that depending on the type, they may contain different classes, Table \ref{table:data_summary} shows the correspondence between the types and their classes. 

\setlength{\tabcolsep}{4pt}
\begin{table}
	\begin{center}
		\caption{Further details of the different types of images in the dataset, based on its industrial processing. This table shows information such as image resolution and the classes present on each type.}
		\label{table:data_summary}
		\begin{tabular}{cccc}
			\hline\noalign{\smallskip}
			Type & \# images & Resolution & Present classes \\
			\noalign{\smallskip}
			\hline
			\noalign{\smallskip}
			Type I & 29 & 1280$\times$959 \& 960$\times$703 & Alpha, TiB2, TiN and FeTiB \\
			Type II & 17 & 1280$\times$959 & Alpha, TiB2 and TiN \\
			Type III & 36 & 1280$\times$959 & Alpha, TiB2, TiN and Fe2B \\
			\hline{\smallskip}
			Total & 82 & 1280$\times$959 \& 960$\times$703 & Alpha, TiB2, TiN, FeTiB and Fe2B \\
			\hline
		\end{tabular}
	\end{center}
\end{table}

\begin{figure}[htbp]
	\caption{Histograms of class distribution per image in the dataset. For each of the three types of images into which the dataset is divided, a graph showing the representation at the level of number of pixels in the whole dataset of each of the classes is displayed. Each color represents a target class: Alpha (purple), TiB2 (red), TiN (yellow), Fe2B (pink), and FeTiB (green).}
	\label{fig:dataset_distribution}
	\centering
	\includegraphics[width=12.5cm, height=10.5cm]{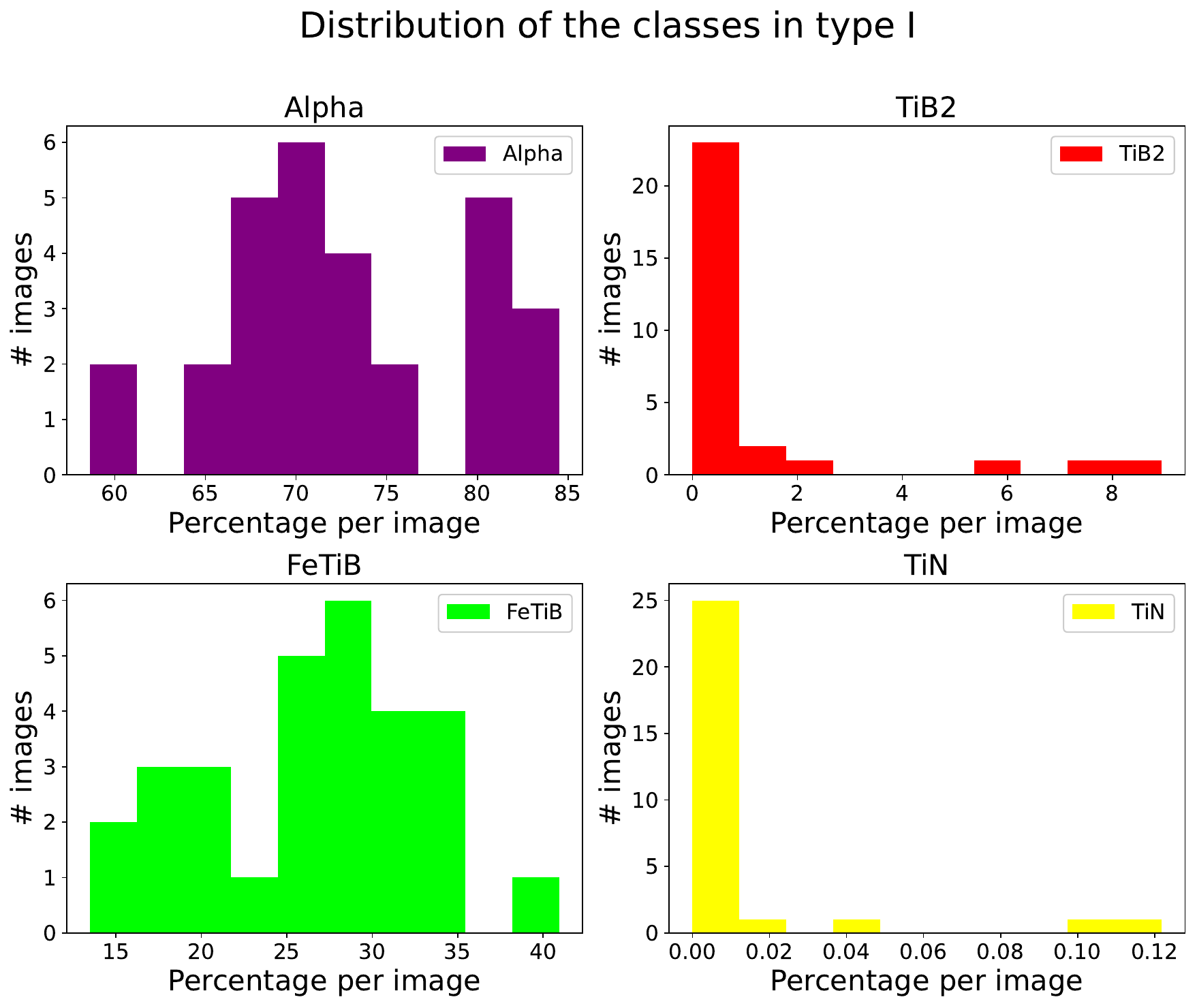}
	\includegraphics[width=13cm, height=5cm]{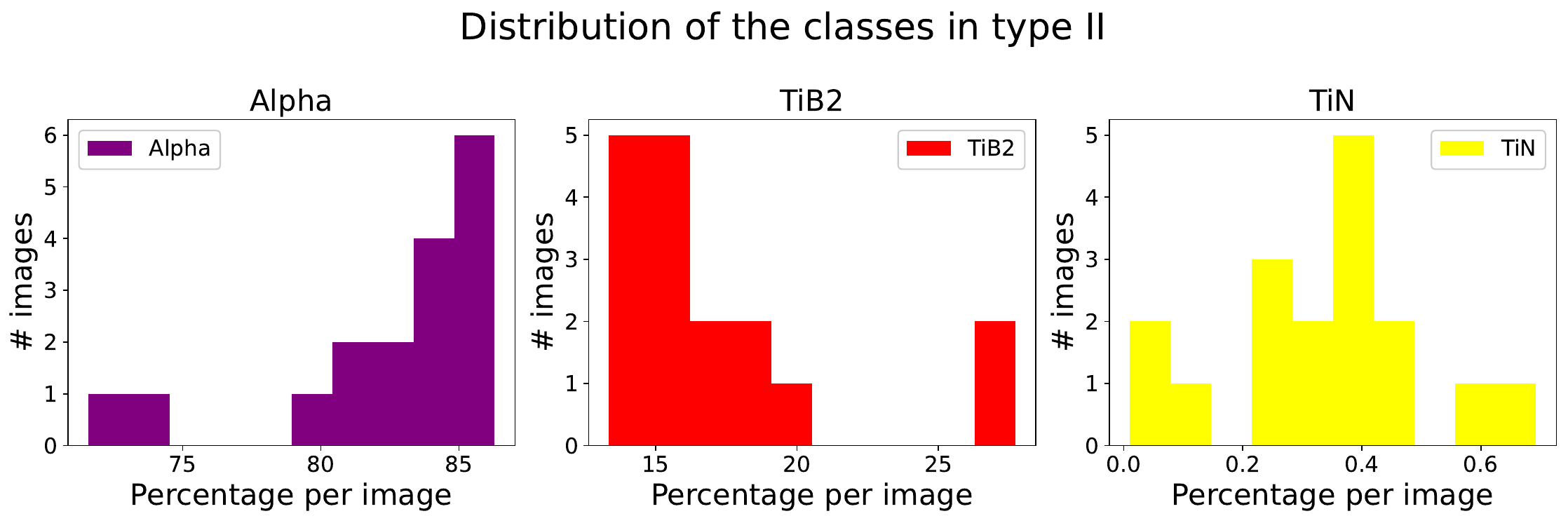}
\end{figure}

Table \ref{table:classes_summary} shows, for each class: the assigned index (from 0 to 4), the color and its percentage in the whole dataset. From this table we notice that it is a highly unbalanced dataset, and, Figure \ref{fig:dataset_distribution} shows a more detailed analysis of the percentage distribution of each class. There are minority classes such as TiN, which are small white dots that form within the TiB2 zones. On the colored label in Figure \ref{fig:dataset_samples}, the TiN zones are represented as small yellow dots always inside the red zones. Additionally, it is worth noting the variation in the distribution of majority classes across different types. For instance, in Type I, the combination of Alpha and FeTiB accounts for 98.99\% of the dataset, while in Type II, Alpha and TiB2 make up 99.67\%. Similarly, in Type III, the majority classes consist of Alpha and Fe2B, which represent 95.12\% of the dataset. This means that, for each type, if we can automate or accelerate the labeling of the 2 majority classes, we will have more than 95\% of the dataset labeled. However, the minority TiN class, together with the TiB2 class, are of greater interest than the other classes from a materials science point of view. Therefore, additional labeling is still required to further enhance the dataset.

\begin{figure}[htbp]
	\centering
	\includegraphics[width=\linewidth]{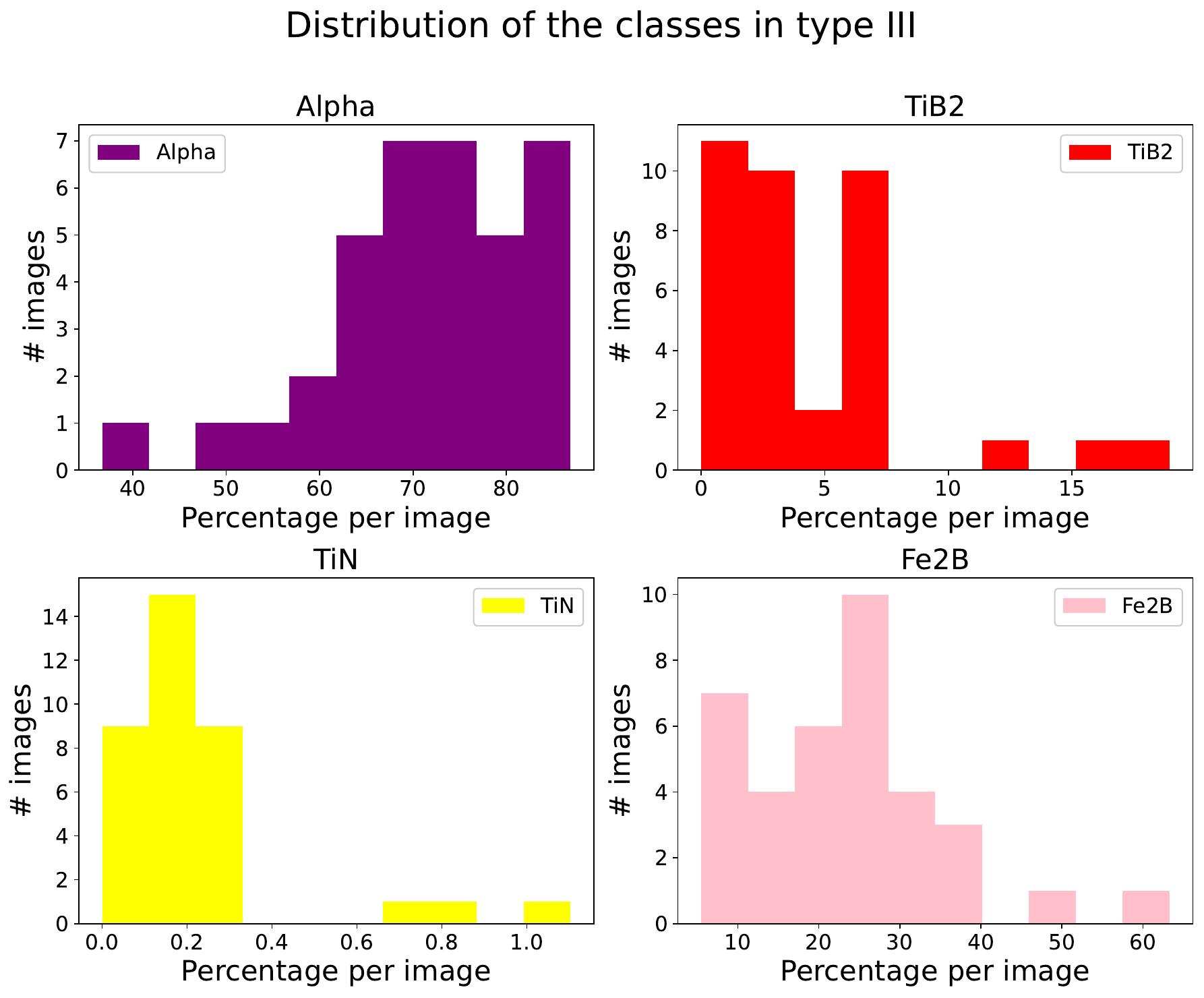}
\end{figure}

\begin{table}[htbp]
\centering
\caption{Average ratio of pixels representing each class in the whole dataset}
\label{table:classes_summary}
\setlength{\tabcolsep}{4pt} 
\begin{tabular}{cccc}
	\hline\noalign{\smallskip}
	Class & Color & Ratio (\%)\\
	\noalign{\smallskip}
	\hline
	\noalign{\smallskip}
	0 - Alpha & purple & 74.22\% \\
	1 - TiB2 & red & 5.97\% \\
	2 - TiN & yellow & 0.17\% \\
	3 - FeTiB & green & 9.22\% \\
	4 - Fe2B & pink & 10.42\% \\
	\hline
\end{tabular}
\end{table}	
	
By dividing the dataset into 3 types despite being the same material, it allows working on individual types or treating the 3 of them globally by training a generic model, which allows testing the generalization capacity of DL models on these datasets.


\section{Experiments}
In this section, we conduct a series of experiments to evaluate and validate the proposed methodology for generating a new dataset on image segmentation. We aim to compare two different labeling strategies and assess the effectiveness of using DL techniques to accelerate the annotation process. Furthermore, we include our experiments with supervised models provide an approximate indication of the performance of DL models on the dataset.

\subsection{Validation of the proposed methodology}
Within this subsection, we validate the effectiveness of the proposed methodology by comparing the two strategies. To do this, three materials-science experts participated in the annotation study. Each image was annotated under both strategies (from scratch and from pre-annotation) in Labelbox, which automatically recorded annotation time. We report the median time across annotators for each image and then aggregate by type. Following standard materials science practice, the dataset was distributed across three experts, each annotating different subsets of images under both strategies. Majority classes (e.g., Alpha, FeTiB/Fe2B) required manual contour tracing rather than technical judgement, while complex minority regions (e.g., TiN inclusions) were marked ``doubtful'' and resolved through consensus meetings among the three experts.

\subsubsection{Pre-annotation analysis}
To proceed with the second strategy, we begin by generating pre-annotations using  5 distinct unsupervised image segmentation techniques. Through quantitative and qualitative analysis, we identify the pre-annotations that closely resemble the actual labels, enabling us to determine the most effective starting point for the annotation process.

\begin{table}[H]
	\begin{center}
		\caption{\textbf{IOU comparison of pre-annotations:} This table shows, divided into each of the three types of images on the dataset, the performance achieved by the  five segmentation algorithms tested as possible pre-annotations. Mean IOU as well as the IOU of each class are given for each of the algorithms.}
		\label{table:pre-annotation_iou}
		\begin{tabular}{ccccccc}
			\hline
			\multicolumn{7}{c}{\bfseries Type I} \\
			\hline\noalign{\smallskip}
			Method & IOU 0 & IOU 1 & IOU 2 & IOU 3 & IOU 4 & Mean IOU \\
			\noalign{\smallskip}
			\hline
			\noalign{\smallskip}
			k-means & \begin{math}67.27\%\end{math} & \begin{math}15.20\%\end{math} & \begin{math}0.00\%\end{math} & \begin{math}36.07\%\end{math} & - & \begin{math}44.06\%\end{math} \\
			multi-Otsu & \begin{math}68.11\%\end{math} & \begin{math}\textbf{16.60}\%\end{math} & \begin{math}0.00\%\end{math} & \begin{math}34.96\%\end{math} & - & \begin{math}44.40\%\end{math} \\
			superpixel & \begin{math}58.91\%\end{math} & \begin{math}10.63\%\end{math} & \begin{math}0.00\%\end{math} & \begin{math}24.87\%\end{math} & - & \begin{math}34.64\%\end{math} \\
			SAM & \begin{math}\textbf{87.07\%}\end{math} & \begin{math}0.17\%\end{math} & \begin{math}0.00\%\end{math} & \begin{math}\textbf{72.06\%}\end{math} & - & \begin{math}\textbf{65.38\%}\end{math} \\
			DL-based & \begin{math}60.89\%\end{math} & \begin{math}7.57\%\end{math} & \begin{math}0.00\%\end{math} & \begin{math}47.18\%\end{math} & - & \begin{math}44.63\%\end{math} \\
			\hline
			\multicolumn{7}{c}{\bfseries Type II} \\
			\hline\noalign{\smallskip}
			Method & IOU 0 & IOU 1 & IOU 2 & IOU 3 & IOU 4 & Mean IOU \\
			\noalign{\smallskip}
			\hline
			\noalign{\smallskip}
			k-means & \begin{math}\textbf{93.66\%}\end{math} & \begin{math}70.83\%\end{math} & \begin{math}0.00\%\end{math} & - & - & \begin{math}\textbf{54.83\%}\end{math} \\
			multi-Otsu & \begin{math}93.56\%\end{math} & \begin{math}70.18\%\end{math} & \begin{math}0.00\%\end{math} & - & - & \begin{math}54.58\%\end{math} \\
			superpixel & \begin{math}87.00\%\end{math} & \begin{math}52.26\%\end{math} & \begin{math}0.00\%\end{math} & - & - & \begin{math}46.42\%\end{math} \\
			SAM & \begin{math}72.30\%\end{math} & \begin{math}25.14\%\end{math} & \begin{math}0.00\%\end{math} & - & - & \begin{math}32.48\%\end{math} \\
			DL-based & \begin{math}70.00\%\end{math} & \begin{math}\textbf{76.89\%}\end{math} & \begin{math}\textbf{0.10\%}\end{math} & - & - & \begin{math}48.84\%\end{math} \\
			\hline\noalign{\smallskip}
			\multicolumn{7}{c}{\bfseries Type III} \\
			\hline\noalign{\smallskip}
			Method & IOU 0 & IOU 1 & IOU 2 & IOU 3 & IOU 4 & Mean IOU \\
			\noalign{\smallskip}
			\hline
			\noalign{\smallskip}
			k-means & \begin{math}\textbf{75.44\%}\end{math} & \begin{math}34.63\%\end{math} & \begin{math}0.00\%\end{math} & - & \begin{math}8.82\%\end{math} & \begin{math}30.05\%\end{math} \\
			multi-Otsu & \begin{math}75.32\%\end{math} & \begin{math}\textbf{35.38\%}\end{math} & \begin{math}0.00\%\end{math} & - & \begin{math}8.07\%\end{math} & \begin{math}30.01\%\end{math} \\
			superpixel & \begin{math}74.23\%\end{math} & \begin{math}18.60\%\end{math} & \begin{math}0.00\%\end{math} & - & \begin{math}14.99\%\end{math} & \begin{math}27.45\%\end{math} \\
			SAM & \begin{math}\textbf{72.30\%}\end{math} & \begin{math}25.14\%\end{math} & \begin{math}0.00\%\end{math} & - & \begin{math}\textbf{0.00\%}\end{math} & \begin{math}\textbf{31.59\%}\end{math} \\
			DL-based & \begin{math}67.11\%\end{math} & \begin{math}18.48\%\end{math} & \begin{math}\textbf{0.04\%}\end{math} & - & \begin{math}\textbf{42.30\%}\end{math} & \begin{math}\textbf{32.40\%}\end{math} \\
			\hline
		\end{tabular}
	\end{center}
\end{table}

Table \ref{tab:pre_metrics} presents Jaccard Index, Dice coefficient, Boundary F1-score, and Hausdorff Distance comparing raw DL-based pre-annotations to final expert labels across Types I-III, confirming strong pre-annotation quality for majority classes while objectively quantifying refinement needs for minority classes and boundaries (Boundary F1 32.6\%, Hausdorff Distance 18.7 pixels).

These metrics demonstrate that DL-based pre-annotations achieve substantial region overlap for dominant microstructure phases (Alpha, FeTiB/Fe2B representing 74--98\% pixels per type), directly enabling the 78\% time savings reported in Table \ref{table:labelling_time}. Lower boundary metrics highlight targeted expert refinement for complex interfaces (TiN inclusions, intricate Alpha/Fe2B boundaries), where manual correction focuses despite comprising $<$6\% total pixels.

\begin{table}[htbp]
	\centering
	\small
	\caption{Raw DL-based pre-annotations quality vs. final labels}
	\label{tab:pre_metrics}
	\begin{tabular}{lcccc}
		\hline
		\textbf{Type} & \textbf{Jaccard} & \textbf{Dice} & \textbf{Boundary F1} & \textbf{Hausdorff distance (px)} \\
		\hline
		Type I   & 44.6\% & 88.4\% & 20.1\% & 12.5 \\
		Type II  & 48.8\% & 77.0\% & 43.9\% & 15.3 \\
		Type III & 32.4\% & 53.8\% & 33.7\% & 22.1 \\
		\hline
		Overall  & 41.5\% & 73.1\% & 32.6\% & 18.7 \\
		\hline
	\end{tabular}
\end{table}

\textbf{Quantitative evaluation:}

This analysis focuses on comparing algorithms performance across the entire dataset rather than selecting optimal methods per individual image. This approach ensures statistically meaningful comparisons while reflecting real-world pre-annotation scenarios where complete ground truth is unavailable during initial labeling. The Intersection over Union (IoU) metrics presented in Table \ref{table:pre-annotation_iou} represent aggregate performance across all images of each material type. This dataset-level analysis illustrates the alignment between pre-annotations and expert-annotated labels.

For \textbf{Type I} images, SAM achieves the highest mean IoU (65.38\%), excelling in the segmentation of the majority classes, Alpha (87.07\%) and FeTiB (72.06\%). The DL-based method demonstrates balanced performance across majority classes compared to traditional methods (k-means, multi-Otsu), despite similar mean IoU values.

In \textbf{Type II} images, k-means outperforms other methods for the Alpha class (93.66\%) and achieves the best mean IoU (54.83\%). In contrast, SAM shows limited effectiveness, with a mean IoU of 32.48\% and particularly poor performance in TiB2 segmentation (25.14\%).

For \textbf{Type III} images, the DL-based method delivers the best overall performance (mean IoU: 32.40\%), especially in Fe2B detection (42.30\%). SAM, however, fails completely in Fe2B segmentation (0.0\%) despite moderate success in Alpha detection (72.30\%).

In summary, the results highlight significant variations in algorithm performance across different image types. SAM excels in Type I images with well-defined objects but struggles in more complex scenarios (Types II and III). This analysis reveals that SAM’s general-purpose architecture, while achieving strong performance on Type I images (65.38\% mean IoU), struggles with the more intricate microstructures of Types II–III (32.48\% and 31.59\% mean IoU, respectively). We suggest that the DL model’s loss, which encourages spatial continuity and groups feature-similar pixels, better captures the morphology of metallographic phases and thereby explains the observed gap [20]. Prompt-engineering variants of SAM could narrow this difference, but they require domain-specific interaction that exceeds the scope of our fully unsupervised setting.

The DL-based method demonstrates consistent reliability across all material types, particularly for critical minority classes like TiB2 and Fe2B. Its ability to detect TiB2 areas is especially valuable from a materials science perspective. Although the DL-based method tends to over-segment the Alpha class, experts prefer this over-segmentation, as it is easier to correct by removing erroneous regions than to manually annotate entire boundaries.

\textbf{Qualitative evaluation:}

\begin{figure}
	\caption{Comparison of (from left to right) original image, pre-annotation and final label for the three MicroSteel types. For each type, the pre-annotation shown is the automatic output of the DL-based method selected by the materials science expert. Class colors are consistent across all panels: Alpha (purple), TiB2 (red), TiN (yellow), Fe2B (pink) and FeTiB (green). This visual comparison illustrates how pre-annotations already capture the majority classes and complex boundaries, requiring only limited expert refinement.}
	\label{fig:preannotations_samples}
	\centering
	a) Type I \\
	\includegraphics[width=3.95cm,height=3.2cm]{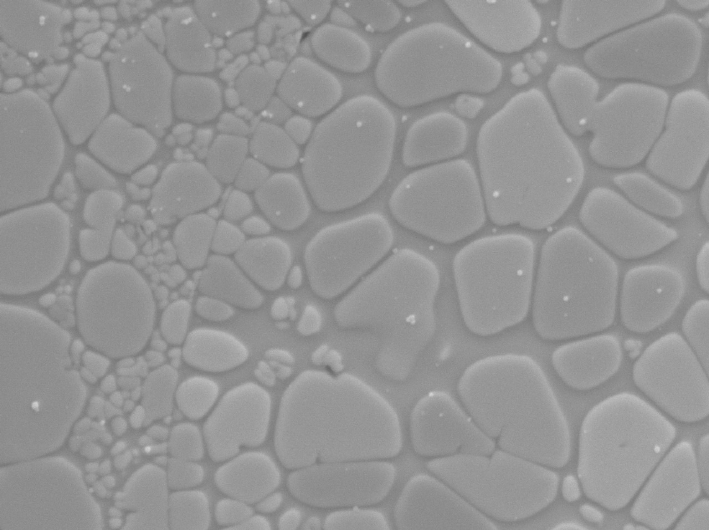}
	\includegraphics[width=3.95cm,height=3.2cm]{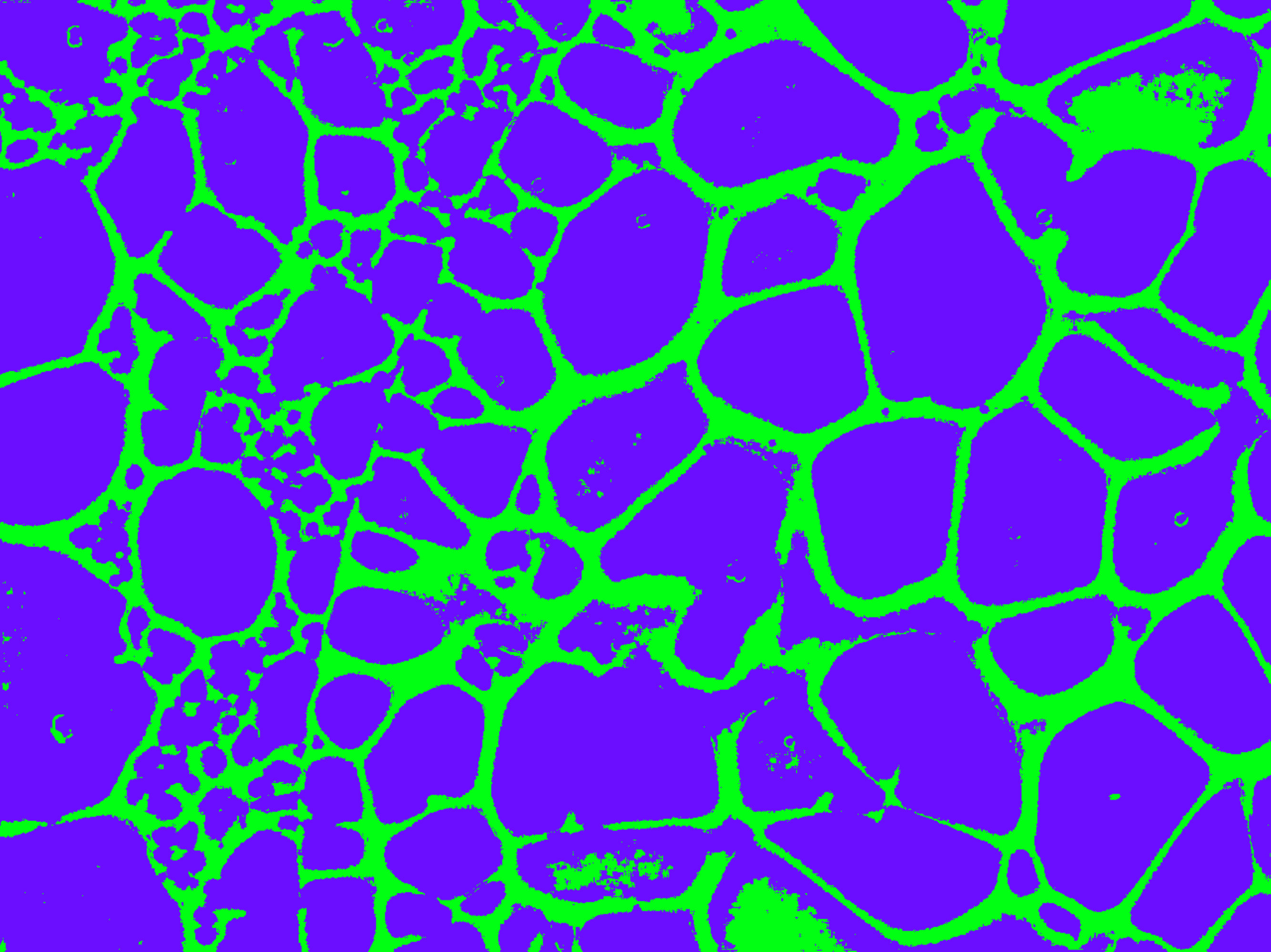}
	\includegraphics[width=3.95cm,height=3.2cm]{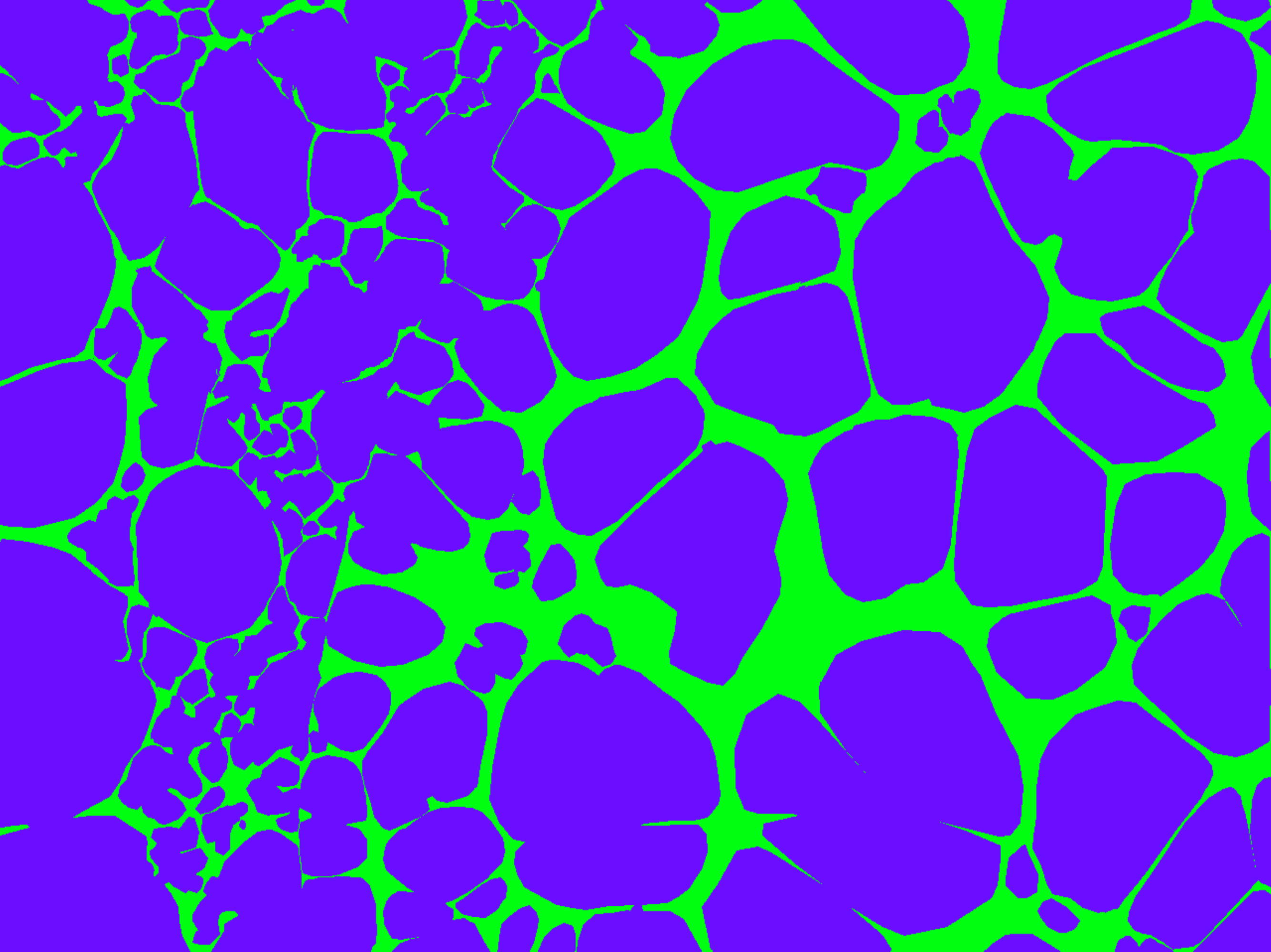}
	\\ b) Type II \\
	\includegraphics[width=3.95cm,height=3.2cm]{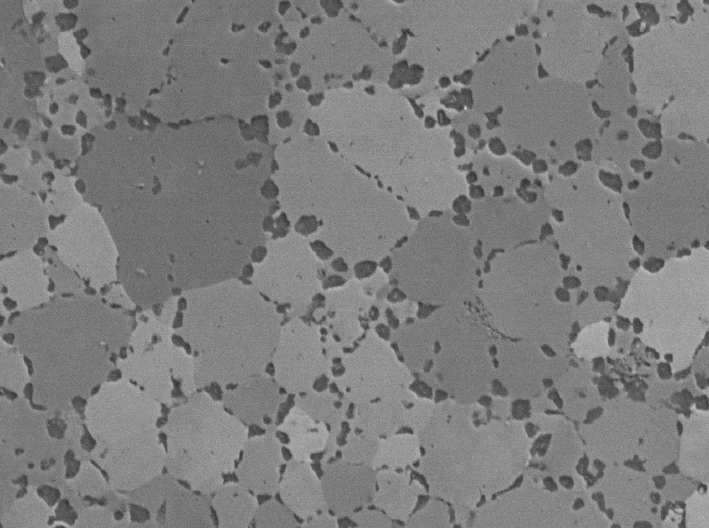}
	\includegraphics[width=3.95cm,height=3.2cm]{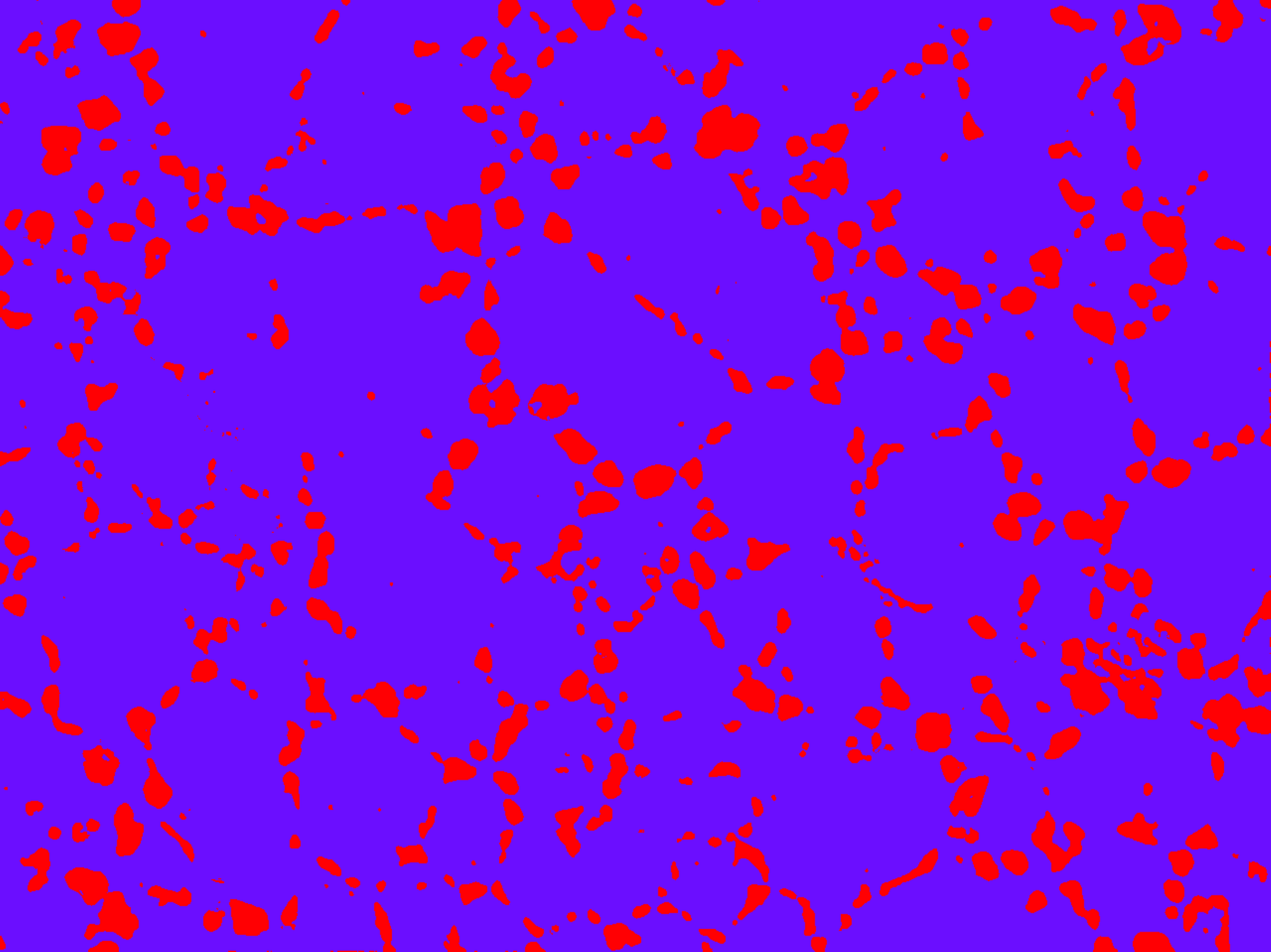}
	\includegraphics[width=3.95cm,height=3.2cm]{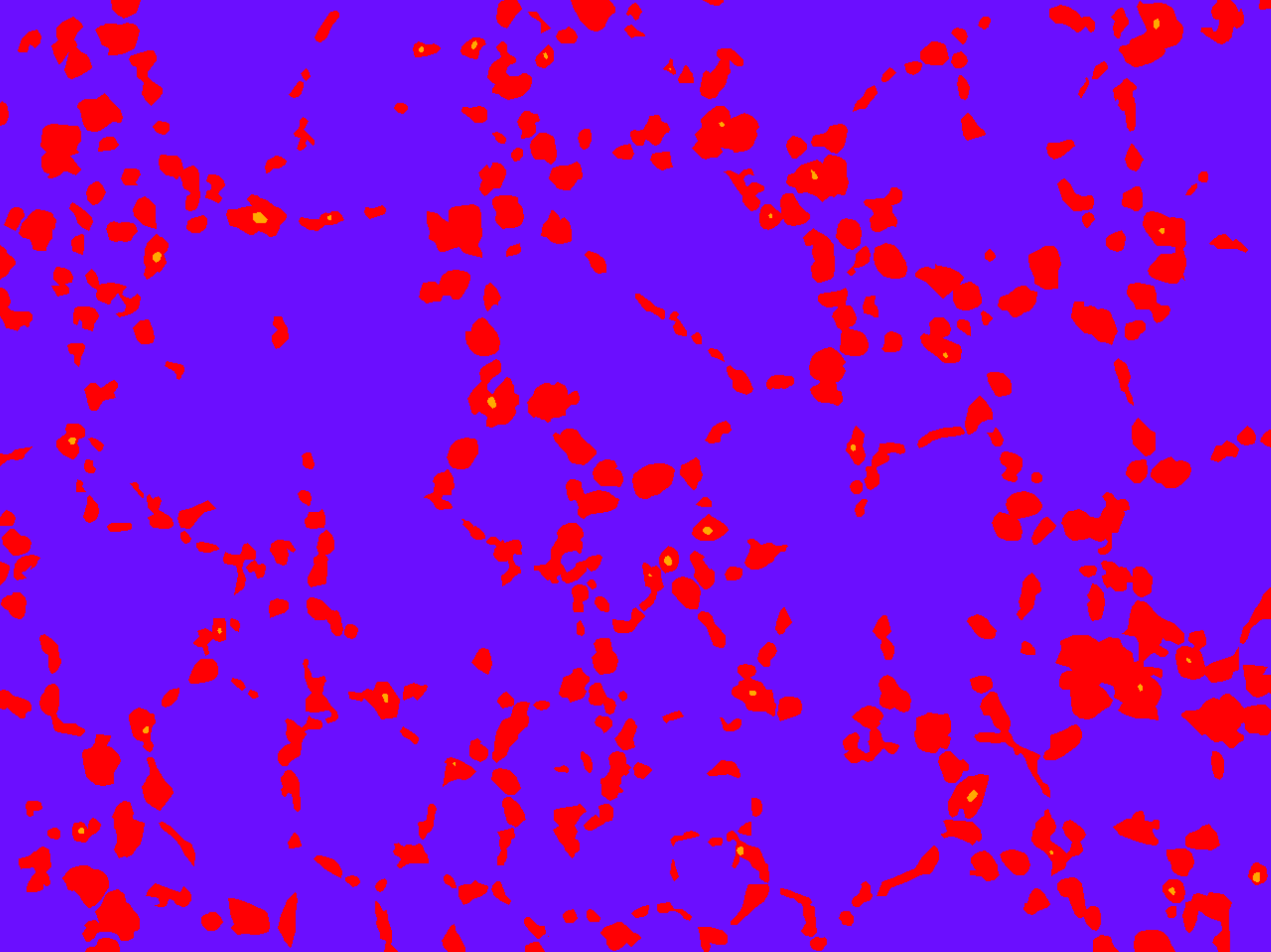}
	\\ c) Type III \\
	\includegraphics[width=3.95cm,height=3.2cm]{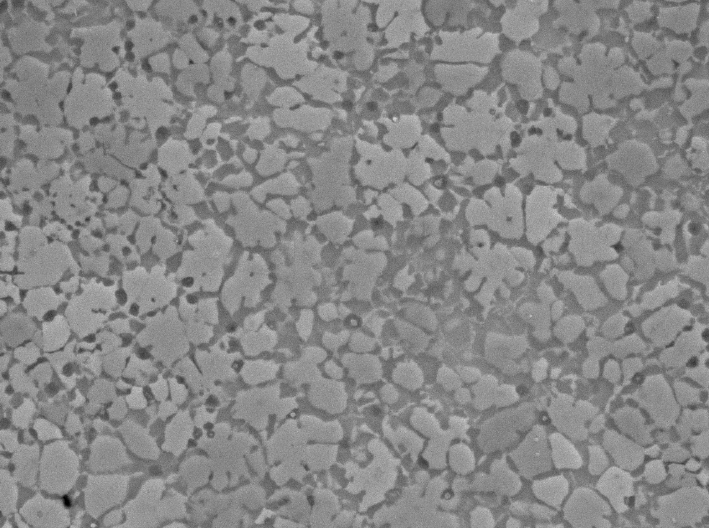}
	\includegraphics[width=3.95cm,height=3.2cm]{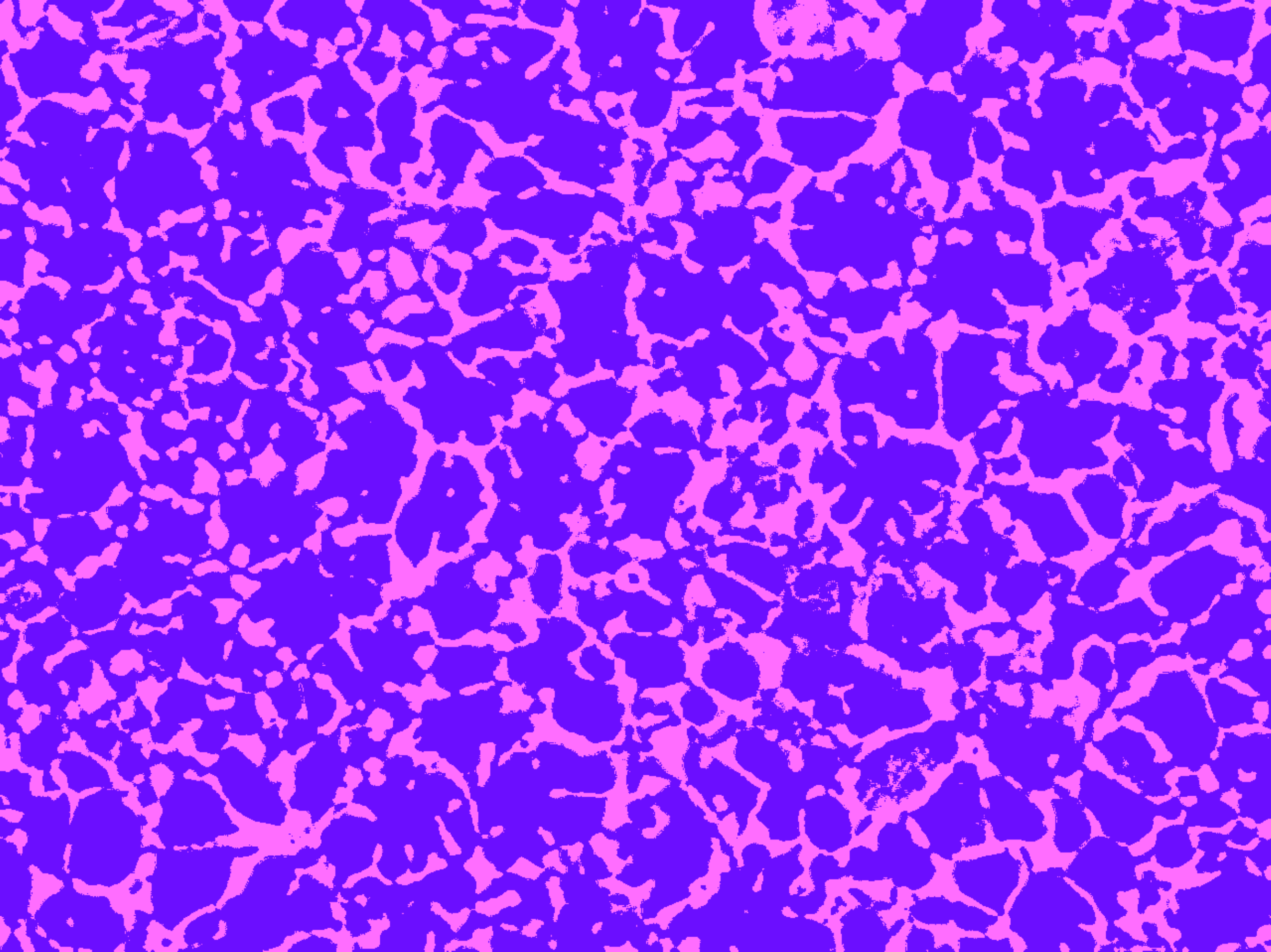}
	\includegraphics[width=3.95cm,height=3.2cm]{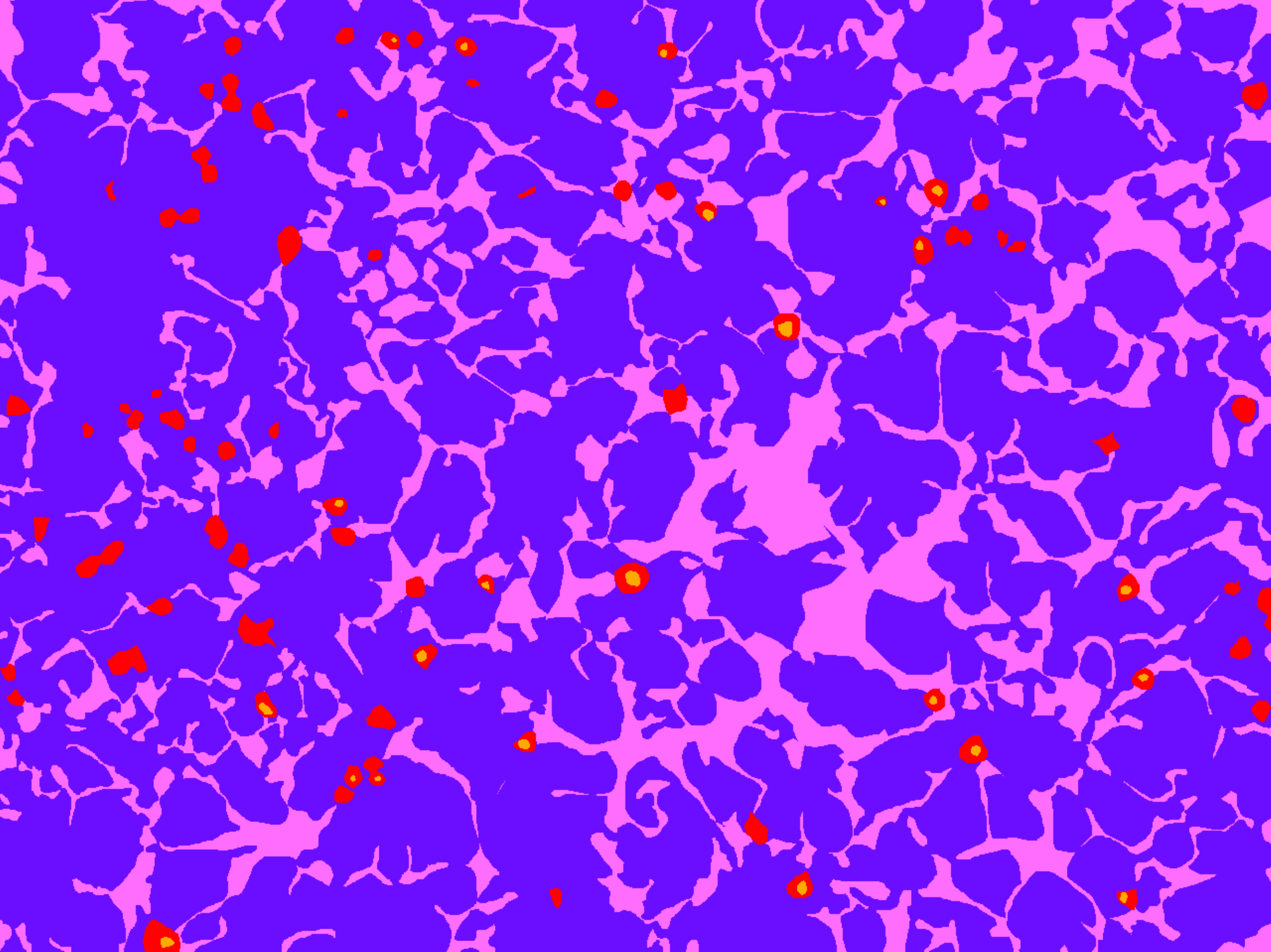}
\end{figure}

Figure \ref{fig:preannotations_samples} presents a visual comparison of the image, pre-annotation generated by the DL-based algorithm, and the final label. The figure demonstrates that the pre-annotations, particularly from the DL-based algorithm, are accurate in types I and II, with only minor discrepancies in minority classes. In Type III, the DL-based approach stands out by accurately segmenting complex boundaries between the major classes Alpha and Fe2B, which experts identified as one of the most challenging aspects of the dataset.

\textbf{Final pre-annotation selection protocol:}
	
Five candidate pre-annotation methods were evaluated (multi-Otsu, superpixels, k-means, SAM, and a DL-based unsupervised method). Their quantitative IOU comparison in Table 5 served as a retrospective validation rather than an operational selection criterion. Although the best-performing method varied by subset (SAM for Type I, k-means for Type II, and DL-based for Type III), the DL-based method was selected as the uniform strategy for the final workflow. This choice was justified by its robustness in the most challenging scenarios (Type III), including superior Fe2B delineation, and its ability to generate boundary structures that minimized manual expert refinement across all image types.

\subsubsection{Annotation quantitative analysis}
In this section, we will compare annotation effort of the two strategies: labeling from scratch and labeling from pre-annotations. For this purpose, the images of the dataset have been labeled with both criteria and their labeling time has been measured for each of the images. Table \ref{table:labelling_time} reports median annotation times across image types, demonstrating 73-84\% time savings with DL pre-annotations. Statistical analysis reveals that pre-annotations not only reduce median time but also dramatically lower inter-image dispersion:

\begin{itemize}
	\item \textbf{Manual scratch annotation:} high variability (std 71.7-99.8 min) across Types I-III
	\item \textbf{DL pre-annotation:} 8-17× lower dispersion (std 1.6-12.3 min), with annotators converging to consistent refinement times
\end{itemize}

\setlength{\tabcolsep}{4pt}
\begin{table}[htbp]
	\centering 
	\caption{\textbf{Labeling time details:} The median time it takes to annotate from scratch and from pre-annotations is compared. For each type, we report the median time per image, total time spent for all images, and the standard deviation of annotation times. * refers to \begin{math}\% \end{math} of time saved by using pre-annotations}
	\label{table:labelling_time}
	\small
	\setlength{\tabcolsep}{2pt}
	\begin{tabular}{l|c|c|c|c|c|c|c}
		\hline
		Type & \multicolumn{2}{c|}{Time per img [min]} &
		\multicolumn{2}{c|}{Total time [h]} &
		\multicolumn{2}{c|}{Standard deviation [min]} &
		Time saved* \\ \cline{2-7}
		& \makecell{From\\scratch} & \makecell{From\\pre-ann.} & \makecell{From\\scratch} & \makecell{From\\pre-ann.} & \makecell{From\\scratch} & \makecell{From\\pre-ann.} & \\ \hline
		I    & 102 & 20 & 58 & 10 & 73.4 &  1.6 & 80.39\% \\ \hline
		II   & 128 & 22 & 36 &  6 & 99.8 & 9.0 & 82.81\% \\ \hline
		III  & 128 & 35 & 76 & 21 & 71.7 & 12.3 & 72.66\% \\ \hline
		\multicolumn{7}{c|}{\bfseries Average time saved} &
		\multicolumn{1}{c}{\bfseries 78\%} \\\hline
	\end{tabular}
\end{table}

The use of pre-annotations substantially reduces the time needed for annotation, with an average time saved of 78\%. This demonstrates that pre-annotations can drastically accelerate the labeling process, making the task more feasible in industrial contexts where time and budget constraints often limit the scope of manual labeling. Specifically, for Type~I, using pre-annotations saves 80.39\% of the time, reducing the total labeling effort from 58~hours to just 10~hours. Type~II sees an even greater time reduction of 82.81\%, while Type~III, although more complex, still achieves a 72.66\% reduction in time. This time reduction for Type~III, despite its complex Fe2B/Alpha boundaries, highlights how even imperfect pre-annotations (mean IoU of 32.40\%) provide a starting point that accelerates annotation.

The efficiency of our approach stems from the DL model's ability to automate majority-class segmentation while preserving structural coherence. By prioritizing spatial continuity in its loss function, the model groups pixels with similar features into contiguous regions, reducing the need for exhaustive boundary adjustments and allowing experts to focus on refining smaller or critical regions.
	
	Notably, the DL method's tendency to over-segment large regions automates the labeling of majority classes, allowing experts to concentrate on critical minority classes:
	\begin{itemize}
		\item The DL method achieves 60.89--76.89\% IoU on majority classes (Alpha, FeTiB, Fe2B), which together cover 74--98\% of pixels (Table \ref{table:classes_summary}), thus automating the largest portion of the annotation work.
		\item Experts then dedicate most of their effort to refine minority classes (TiB2, TiN), which occupy less than 6\% of pixels.
		\item Pre-annotations make labeling less tedious: experts refine TiB2/TiN more easily by iterating over multiple small objects, rather than spending most of the time drawing thin, complex boundaries between majority classes.
\end{itemize}

From these results, it is evident that annotating the entire dataset from scratch would require approximately 6~days and 18~hours of non-stop work, whereas using DL-based pre-annotations reduces this to just 1~day and 13~hours. This significant reduction not only highlights the practical advantages of DL techniques in materials science applications but also demonstrates their potential to make otherwise infeasible projects viable.

\subsection{Model development for industrial application}
Finally, we present the results obtained by fully supervised DL models trained on each type of material, with their test results verified by materials science experts. These models have proven to be effective, leading to their deployment for industrial purposes. It is important to note that the main scope of this article is the labeling, analysis, and dataset generation process, so the examination of various network architectures and hyperparameter optimization falls beyond the scope of this study.

We have used PyTorch Segmentation Models library \footnote{\url{https://smp.readthedocs.io/en/latest/}} for training four models with a DeepLabV3+ \cite{chen2018encoder} with EfficientNet-b0 backbone \cite{tan2019efficientnet}. This selection was based on its proven success in similar cases within the field of materials science. One of the four models has been trained on all three types, considering them collectively as a unified set. It was subsequently tested on each type individually, and its corresponding results are depicted in Table \ref{table:results_supervised} as \emph{G-X}, where X denotes the type index. On the other hand, the remaining 3 models were trained and tested exclusively on individual types, denoted as \emph{S-X}. All supervised models were trained for a fixed number of epochs (200 epochs with a patience of 30 epochs for early stopping), which was sufficient to ensure convergence on all dataset splits, as verified by expert validation of the resulting segmentations.

The results presented in Table \ref{table:results_supervised} reveal several key insights into the performance and applicability of supervised models for microstructure segmentation:

\begin{itemize}
		\item \textbf{Type-Specific Models Outperform General Models}: The \emph{S-X} models consistently achieve higher mean IoU than the \emph{G-X} models. For example, the S-III model achieves 67.14\% mean IoU compared to 54.81\% for G-III. This highlights the significance of distinguishing between the three subsets of types in this dataset, as each type have different phase distributions and boundary complexities. Models trained on distinct subsets learn phase morphology differences, and the general model struggles with inter-type variations, particularly in distinguishing visually similar classes like FeTiB (Class 3) and Fe2B (Class 4).
		
		\item \textbf{Importance of majority classes}: The models are specially good in segmenting majority classes (like Alpha, FeTiB, Fe2B), with higher IoU scores.
		
		\item \textbf{Challenges of minority classes}: The models achieve relatively lower results for minority classes, particularly TiN (0.00–37.90\% IoU). This is due to several factors:
		\begin{itemize}
			\item \textbf{Class imbalance}: TiN occupies only 0.17\% of pixels (Table \ref{table:classes_summary}), making it statistically insignificant during training, despite loss penalization being adjusted based on class distribution
			\item \textbf{Noisy labels}: TiN particles often blend with TiB2 in optical microscopy, leading to inconsistent annotations even among experts.
		\end{itemize}
\end{itemize}

These results demonstrate that the DL-based pre-annotation strategy not only accelerates dataset generation but also enables the development of robust supervised models tailored to industrial needs. By automating majority-class labeling and providing a consistent foundation for supervised training, the DL method ensures that experts can focus on refining critical minority classes, ultimately leading to more accurate and reliable models.

\setlength{\tabcolsep}{4pt}
\begin{table}[H]
	\begin{center}
		\caption{\textbf{Results of supervised models:} \emph{G*-X**} approaches refer to the general model trained over all data and evaluated on type \emph{X}, \emph{S***-X} approaches are trained and tested on a single type \emph{X}. *G refers to the model tested on all three types of images. ** X Denotes the index of the type of image subset. ***S refers to the model trained and tested exclusively on individual types.}
		\label{table:results_supervised}
		\begin{tabular}{ccccccc}
			\hline\noalign{\smallskip}
			Model & IOU 0 & IOU 1 & IOU 2 & IOU 3 & IOU 4 & Mean IOU \\
			\noalign{\smallskip}
			\hline
			\noalign{\smallskip}
			G &
			\begin{math}92.41\%\end{math} & \begin{math}62.85\%\end{math} & \begin{math}26.91\%\end{math} & \begin{math}72.65\%\end{math} & \begin{math}76.03\%\end{math} & \begin{math}66.17\%\end{math} \\
			\hline
			G-I & 
			\begin{math}93.93\%\end{math} & \begin{math}\textbf{31.74\%}\end{math} & \begin{math}\textbf{3.39\%}\end{math} & \begin{math}73.31\%\end{math} & - & \begin{math}40.37\%\end{math} \\
			S-I &
			\begin{math}\textbf{94.31\%}\end{math} & \begin{math}18.15\%\end{math} & \begin{math}0.00\%\end{math} & \begin{math}\textbf{81.11\%}\end{math} & - & \begin{math}\textbf{48.39\%}\end{math} \\
			\hline
			G-II &
			\begin{math}91.99\%\end{math} & \begin{math}63.14\%\end{math} & \begin{math}19.40\%\end{math} & - & - & \begin{math}34.91\%\end{math} \\
			S-II &
			\begin{math}\textbf{92.25\%}\end{math} & \begin{math}\textbf{66.28\%}\end{math} & \begin{math}\textbf{37.90\%}\end{math} & - & - & \begin{math}\textbf{65.48\%}\end{math} \\
			\hline
			G-III & 
			\begin{math}\textbf{91.22\%}\end{math} & \begin{math}\textbf{67.27\%}\end{math} & \begin{math}33.25\%\end{math} & - & \begin{math}\textbf{82.31\%}\end{math} & \begin{math}54.81\%\end{math} \\
			S-III & 
			\begin{math}90.54\%\end{math} & \begin{math}61.59\%\end{math} & \begin{math}\textbf{36.37\%}\end{math} & - & \begin{math}80.07\%\end{math} & \begin{math}\textbf{67.14\%}\end{math} \\
			\hline
		\end{tabular}
	\end{center}
\end{table}

\section{Conclusions}
In this paper, we have addressed the challenge of accelerating the data annotation process for industrial semantic segmentation applications. We have quantified and explored ways to speed up the laborious task of manual data annotation, which can be a significant bottleneck in real-world projects, particularly in industrial settings such as materials characterization.

By leveraging unsupervised deep neural networks, we have demonstrated significant reductions in the time and effort required for data labeling. Our study has shown that using a deep model for pre-annotations can reduce the time to label a complex industrial dataset from 6 days and 18 hours to 1 day and 13 hours, with a mean percentage reduction of approximately \begin{math}78\%\end{math}. These results are crucial in scenarios where time or budget constraints play a significant role in project decision-making.

Furthermore, we share the new real-world dataset MicroSteel under MIT License with permanent archival on Zenodo [DOI: 10.5281/zenodo.18826160]: to the best of our knowledge, the largest existing dataset in the field of microstructure characterization. The proposed annotation process enables the generation of larger volumes of annotations and facilitates the development of semantic segmentation models. We also provide metrics on the test set for both unsupervised and supervised models validated by an expert in the field and deployed in production.

\section{\textbf{Limitations and Future Work}}

While our methods have shown significant improvements in annotation efficiency, there are some limitations to consider. While our study demonstrates significant benefits of pre-annotation in accelerating the annotation process for metallographic images (an average of ~78\% time reduction, as shown in Table \ref{table:labelling_time}), it is important to acknowledge that the effectiveness can be context-dependent. However, for the specific metallography problem addressed, expert feedback consistently indicated that refining pre-annotations, particularly for prevalent majority classes (e.g., Alpha, FeTiB), was considerably less burdensome than starting from scratch. This allowed experts to efficiently dedicate their focus to critical, often minority, classes (e.g., TiN), confirming a net benefit in annotation efficiency for this domain.

Although the proposed annotation workflow can potentially benefit other semantic-segmentation problems characterized by high-resolution images, class imbalance, and large homogeneous regions, the quantitative speedup is expected to vary across domains. In the present dataset, gains were consistent across all three subsets (Type I: 80.39\%, Type II: 82.81\%, Type III: 72.66\%; average: 78\%), supporting internal robustness. In addition, prior cross-domain evidence from MetalDAM, CamVid, FloodNet, and NuCLS showed meaningful unsupervised pre-annotation quality with 0\% manually labeled pixels (mean IoU range: 19.30\%–32.24\%) \cite{fernandez2023tradeoff}, providing preliminary evidence that the pre-annotation principle can recover useful structure across domains. However, as is intrinsic to machine-learning systems, transfer performance and operational gains depend on data distribution, class structure, and annotation protocol. Accordingly, the 78\% reduction should be interpreted as representative of this case study rather than universally fixed.

Furthermore, it is critical to address potential failure cases where pre-annotation could hinder the workflow rather than accelerate it. In our protocol, the primary qualitative criterion for selecting the DL-based method over classical alternatives was its capacity to generate highly continuous contours, avoiding the fragmented or 'granular' over-segmentation that requires tedious pixel-by-pixel corrections. By successfully automating the definition of complex main boundaries between unstructured phases, the most demanding technical effort is mitigated, leaving experts with the straightforward task of introducing thicker, linear features that do not demand highly meticulous refinement. If an optimized algorithm suffers from severe over-segmentation or hallucinates incorrect boundaries, a phenomenon of negative transfer occurs. In such failure scenarios, the mechanical effort required to fix the initial mask can ultimately exceed the time required to annotate the raw image from scratch.

In this context, while foundation models such as SAM can potentially benefit from prompt engineering, such techniques require domain-specific interaction that contradicts our objective of minimizing human intervention and taking as much advantage as possible of fully unsupervised pre-annotations. Future work could therefore explore hybrid strategies that leverage recent advances in foundation and multimodal models while preserving annotation efficiency. In particular, newer generations of segmentation foundation models (e.g., SAM-2 and emerging SAM-3-style architectures) could be evaluated as pre-annotation engines, especially in challenging cases with ambiguous or fragmented boundaries. In addition, vision-language models (e.g., CLIP-based approaches) could provide high-level semantic cues that help constrain segmentation, for instance by guiding region selection or filtering candidate masks. An important open direction is to study the trade-off between additional human interaction (prompts) and further reductions in expert refinement time, and to design weakly supervised or fully automatic variants that retain the scalability of our workflow.

In conclusion, our work provides valuable insights and practical solutions for accelerating the annotation process in industrial semantic segmentation applications, with implications beyond the specific use case of materials characterization.

\section{Acknowledgements}
The authors acknowledge Valérie Daeschler and Frédéric Bonnet in ArcelorMittal Global R\&D Maizières for acquiring and providing the micrographs for the MicroSteel dataset. Funding for open access charge: Universidad de Granada / CBUA. This work was supported by grant CONFIA (PID2021-122916NB-I00) funded by MCIN/AEI/ 10.13039/501100011033 ,  and by grant FORAGE (B-TIC-456-UGR20) funded by Consejería de Universidad, Investigación e Innovación, both funded by ``ERDF A way of making Europe"

\bibliography{mybibfile}
	
\end{document}